\documentclass{article} % For LaTeX2e
\usepackage{nips15submit_e,times}
\usepackage[numbers]{natbib}

\usepackage{amsmath,amsfonts,bm}

\def\eqref#1{equation~\ref{#1}}
\def\1{\bm{1}}

\DeclareMathAlphabet{\mathsfit}{\encodingdefault}{\sfdefault}{m}{sl}
\SetMathAlphabet{\mathsfit}{bold}{\encodingdefault}{\sfdefault}{bx}{n}

\usepackage{url}
\usepackage{graphicx}
\usepackage{booktabs}
\usepackage{amsmath,amssymb}
\usepackage{array}
\usepackage{multirow}
\usepackage{float}
\usepackage{hyperref}
\hypersetup{
  hidelinks,
  pdftitle={Residual Algebra for Representation-Preserving Learning},
  pdfauthor={Yao Wu}
}

\newtheorem{proposition}{Proposition}[section]

\newcommand{\BPQ}{B_{\mathrm{PQ}}}
\newcommand{\Xpar}{X_{\mathrm{par}}}
\newcommand{\pp}{\,\mathrm{pp}}
\newcommand{\Rop}{\mathcal{R}_{\mathrm{PQ}}}
\newcommand{\Ropm}{\mathcal{R}_{\mathrm{PQ},m}}
\newcommand{\Aop}{\mathcal{A}}
\newcommand{\Cop}{\mathcal{C}}
\newcommand{\Fvec}{\mathbf{F}}

\title{Residual Algebra for\\Representation-Preserving Learning}

\author{
Yao Wu \\
\texttt{wuyao@westlake.edu.cn}
}

\nipsfinalcopy
\begin{document}

\maketitle

\begin{abstract}
Learning from heterogeneous representations is often reduced to feature concatenation, erasing which representation produced each error. We propose \emph{residual algebra}, in which each representation retains its coordinate system and owns its unresolved residual until an explicit aggregation boundary. \textbf{Fold} instantiates representations as point-in-time conditional-mean fields on $10{\times}10$ rank grids, and \textbf{FPRC-PQ} composes them through \emph{relax--aggregate--close}: each field first fits a correction to its own residual, corrected fields then meet at a fixed mean, and a shared learner closes only the aggregate's fresh residual. We formalize aggregation as a quotient by the zero-sum redistribution kernel, characterizing legal post-aggregation operators as those constant on its cosets. The resulting composition separates representation, local residual estimation, and residual-of-residual estimation, with population variance reduction and first-order coupled-path mean orthogonality. Rumination-B and Rumination-H extend the algebra with quotient-legal finite correction and feedback. On 3.67M Chinese A-share stock-day observations (2023--2026) under a frozen point-in-time protocol, FPRC-PQ raises net-of-cost return from 13.52\% to 19.10\% and Sharpe from 1.42 to 2.09, outperforming matched-capacity, unified-residual, identity-free two-stage, and pairwise-only controls. The gain is thus attributable to explicit residual ownership and composition rather than additional features or trees.
\end{abstract}

\section{Introduction}
\label{sec:intro}

Cross-sectional return prediction is a canonical hard tabular problem: several thousand entities are re-ranked daily from a few dozen noisy covariates whose joint distribution drifts continuously \citep{gu2020empirical,cont2001empirical,gama2014survey}. On such data, gradient-boosted decision trees remain a reference model class \citep{chen2016xgboost,grinsztajn2022tree,mcelfresh2023neural,shwartzziv2022tabular}, while deep tabular alternatives often only match them \citep{gorishniy2021revisiting,hollmann2025tabpfn,borisov2024deep}. Yet both are commonly asked to learn from one flat feature vector. Concatenation preserves values but destroys provenance: after heterogeneous representations are pooled, the learner no longer knows which representation generated which error. The central question of this paper is therefore:

\textbf{Core question (residual algebra).} Can representation-specific errors be made composable without first erasing the identity of the representation that owns them? Simply growing one learner conflates heterogeneous error populations; simply averaging experts discards their identities \citep{jacobs1991adaptive,wolpert1992stacked}.

Answering it requires two ingredients: explicit representation objects and operators with controlled type boundaries.

\textbf{Objects.} An interaction must become a stable object with its own coordinates and residual, rather than an implicit collection of tree paths.

\textbf{Operators.} Local correction must act before identity erasure; aggregation must expose an explicit boundary; and any shared correction must target only the fresh residual left by the preceding operators.

We call this construction \emph{residual algebra}: residuals are typed by their generating representations, and valid learning stages are defined by their domains, targets, and composition order.

\textbf{Representation layer---Fold (\S\ref{sec:fold}).} Fold maps a factor pair to a $10{\times}10$ grid of daily rank deciles. Each cell is a discrete interaction state, and its point-in-time conditional mean defines a factor field $F_m$. In the population, $F_m$ is a projection onto the information carried by that grid; $r-F_m$ is therefore exactly what the representation leaves unresolved. Three heterogeneous fields provide distinct local coordinate systems and three typed residual populations.

\textbf{Residual algebra---FPRC-PQ (\S\ref{sec:fprc}).} Field-Preserving Residual Closure instantiates three ordered operators. Relaxation maps every $(F_m,r-F_m)$ to a corrected but still typed field $H_m$; aggregation maps the product of typed fields to one shared backbone $B$ and is the only identity-erasure step; closure adds a model of $r-B$ without rereading local state. The aggregation boundary is a quotient: it identifies exactly the typed configurations that differ by zero-sum field redistribution, and every legal downstream map must factor through that quotient. The algebra also exposes exact telescoping, a control-variate correction, and a coupled-path orthogonality of the final mean. Its value is structural rather than terminological: changing the residual owner, operator order, information boundary, or capacity placement produces a different model that can be tested directly.

\textbf{Analytical extensions---two rumination loci (\S\S\ref{sec:rumination}--\ref{sec:rumination-h}).} At the aggregate backbone, typed routes end, shared routes begin, and erased identity cannot return. Rumination-B performs one finite algebraic correction inside the legal quotient space. Rumination-H instead treats residual-born discrepancy as feedback: it reads from the aggregate state but writes to the fixed residual anchor before typed relaxation is rerun. Both are mechanism-level extensions, not additional result rows in the present evaluation.

\textbf{Evidence discipline.} All architectural choices, factor selections, and hyperparameters were frozen on data through 2022; evaluation covers 2023--2026 under annual point-in-time refits, realistic costs, and paired 21-day block-bootstrap inference \citep{kunsch1989jackknife,lopez2018advances}. Five operator audits and four matched structural controls isolate residual ownership from total tree capacity, pairwise interaction class, and generic staging. Eight extra meta-information channels also fail, several significantly, showing that side information cannot repair an incorrectly placed correction stage.

Our contribution is a representation-learning principle with an executable form: \textbf{algebraize the residual before optimizing it}. Fold supplies explicit representation objects; FPRC supplies typed operators and a quotient boundary theorem for identity erasure; Rumination-B allocates the legal post-aggregation innovation space; Rumination-H turns quotient-compatible diagnosis into identity-neutral reflux and typed reinterpretation; and the frozen audits test neighboring base algebras rather than presenting only a winning pipeline. The resulting account separates why the system works from which market variables instantiate it.

\section{Fold: from continuous factor pairs to interaction states}
\label{sec:fold}

\begin{figure}[t]
\centering
\includegraphics[width=\linewidth]{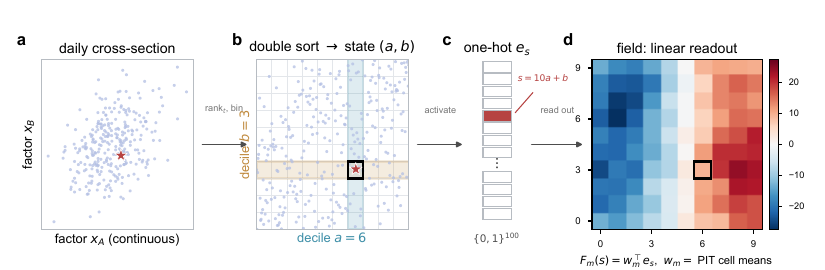}
\caption{\textbf{Fold is a double sort read as a one-hot encoding.} \textbf{(a)}~A daily cross-section in continuous factor space; one stock highlighted. \textbf{(b)}~Both factors are ranked within the day and cut into deciles; the rank pair selects exactly one of $10{\times}10$ interaction states (bands: the stock's column decile $a{=}6$ and row decile $b{=}3$). \textbf{(c)}~The induced indicator $e_s\in\{0,1\}^{100}$ has a single active entry at $s=10a+b$. \textbf{(d)}~The field is the linear readout $F_m(s)=w_m^{\top}e_s$ whose weight table holds PIT cell means of forward returns (surface illustrative).}
\label{fig:fold}
\end{figure}

\textbf{Construction.} Let $\mathcal{U}_t$ be the tradable universe on day $t$ and let $x_{A,i,t},x_{B,i,t}$ be two continuous factors. Within each day, both factors are rank-transformed and binned into deciles,
\begin{equation}
a_{i,t}=\big\lceil 10\,\mathrm{rankpct}_t(x_{A,i,t})\big\rceil-1,\qquad
b_{i,t}=\big\lceil 10\,\mathrm{rankpct}_t(x_{B,i,t})\big\rceil-1,
\end{equation}
yielding a discrete \emph{interaction state} $s_{i,t}=(a_{i,t},b_{i,t})\in\{0,\dots,9\}^2$. A \emph{factor field} is the point-in-time conditional-mean map
\begin{equation}
F_m(s)=\widehat{\mathbb{E}}_{\mathrm{PIT}}\big[\,r_{i,t}\,\big|\,s_{i,t}=s\,\big],
\label{eq:field}
\end{equation}
where $r_{i,t}$ is the 5-day forward return and the estimate uses only history observable at deployment time (annual refresh, sample weights with a 252-day half-life). Equivalently, Fold is a \emph{double sort read as a one-hot encoding} (Fig.~\ref{fig:fold}): with $e_s\in\{0,1\}^{100}$ the indicator of the active cell, the field is the linear readout $F_m(s)=w_m^{\top}e_s$ whose weight table holds the PIT cell means. The interaction is therefore a first-class object rather than a path carved implicitly through a tree ensemble. With ${\sim}4{,}000$ stocks per day, every cell receives ${\sim}40$ fresh observations daily, so \eqref{eq:field} is dense and statistically stable while remaining a $100$-entry table that can be plotted, audited, and diffed across years.

Three design properties matter downstream. \emph{(P1) Regime robustness:} states are within-day order statistics, invariant to level and scale shifts of the raw factors \citep{cont2001empirical}. \emph{(P2) Interpretability:} each cell is a human-readable conjunction of two ordinal states, and $F_m$ is an explicit interaction surface rather than a path through a tree ensemble \citep{lou2013accurate,caruana2015intelligible}. \emph{(P3) Portability:} the construction consumes any ordered feature pair; nothing in the residual algebra of \S\ref{sec:fprc} references financial semantics. The grid is therefore a concrete carrier of representation identity, not an assumption about a particular factor family.

\textbf{Heterogeneous fields.} We instantiate $M{=}3$ fields from distinct, preselected economic mechanisms, denoted only by $F_1,F_2,F_3$. Their signals correlate only moderately (mean pairwise $0.54$), which is precisely why identity preservation---not blending---becomes the central architectural question. The exact factor identities and construction transforms are commercially sensitive and are not reported.

\textbf{Auxiliary coordinates.} Local correction also receives a fixed auxiliary interface $Z_{\mathrm{aux}}=(P,Q)$ shared across experts. We deliberately abstract away the economic identities, dimensionality, preprocessing, and screening statistics of these coordinates. They were fixed before the evaluation window and are unchanged across all reported structural controls. This separation keeps the paper's object explicit: the claims concern residual ownership, operator order, and information routing conditional on one frozen input interface, rather than disclosure of a tradable signal recipe.

\textbf{Projection view: what a field leaves unresolved.} Let $\mathcal{G}_m=\sigma(s_m)$ be the information carried by field $m$'s grid and let $\mathcal{K}_m=L^2(\mathcal{G}_m)$, equivalently the span of its 100 cell indicators inside $\mathcal{H}=L^2(\Omega,\mathcal{F},\mathbb{P})$. The population field is the orthogonal projection $\Pi_m r=\mathbb{E}[r\mid\mathcal{G}_m]$; the PIT table $F_m$ is its finite, shrinkage-regularized estimate. Hence the population residual $(I-\Pi_m)r$ is orthogonal to every function of the grid state alone. A useful correction must therefore enlarge the local information set, here to $\mathcal{L}_m=L^2\!\left(\sigma(s_m,Z_{\mathrm{aux}})\right)\supset\mathcal{K}_m$, while retaining which $\mathcal{K}_m$ produced the residual. Because the field subspaces overlap and are not mutually orthogonal, their mean is not a joint projection; it is an explicit interface across which identity is deliberately erased. This projection-to-residual transition motivates the architecture that follows.

\section{Residual algebra: field-preserving relaxation and closure}
\label{sec:fprc}

The evaluated instance, FPRC-PQ, seats the fixed auxiliary interface $Z_{\mathrm{aux}}$ inside each field's local correction. Two naive extremes bracket its design space. Pooling the fields, $\bar F=\frac1M\sum_m F_m$, discards which field produced which opinion and earns 13.5\% net over the evaluation window (Table~\ref{tab:ladder}). A single shared corrector on top of $\bar F$ treats three heterogeneous residual populations as one and reaches 18.0\%. FPRC-PQ instead makes the residual an algebraic object and composes three learned nonlinear operators---a field-preserving \emph{relaxation} $\Rop$, an identity-erasing \emph{aggregation} interface $\Aop$, and a fresh-residual \emph{closure} $\Cop$ (Fig.~\ref{fig:fprc}). Their type signatures are
\begin{equation}
\begin{aligned}
\Rop &: \prod_{m=1}^{M}\mathcal K_m\longrightarrow\prod_{m=1}^{M}\mathcal L_m,
&\Aop &: \prod_{m=1}^{M}\mathcal L_m\longrightarrow\mathcal H,\\
\Cop &: \mathcal H\longrightarrow\mathcal H.
\end{aligned}
\label{eq:operator-types}
\end{equation}
The product type before $\Aop$ retains the field index; the scalar backbone after $\Aop$ does not. This is the controlled identity-erasure boundary. Composition is therefore part of the model definition, not notation that can be reordered freely:
\begin{equation}
\boxed{\ \widehat{S}=\big(\Cop\circ\Aop\circ\Rop\big)(\Fvec),\qquad \Fvec=(F_1,\dots,F_M).\ }
\label{eq:composition}
\end{equation}
Expanding the operators makes their residual targets, information domains, and routing boundaries explicit:
\begin{align}
\text{(relax)}\quad &\Ropm(F_m) \;=\; F_m + g_m\big(a_m,\,b_m,\,Z_{\mathrm{aux}};\ \ r-F_m\big) \;=:\; H_m, \label{eq:local}\\
\text{(aggregate)}\quad &\Aop(H_1,\dots,H_M) \;=\; \tfrac{1}{M}\textstyle\sum_{m=1}^{M} H_m \;=:\; \BPQ, \label{eq:mean}\\
\text{(close)}\quad &\Cop(\BPQ) \;=\; \BPQ + G\big(\Xpar;\ \ r-\BPQ\big) \;=:\; \widehat{S}, \label{eq:closure}
\end{align}
where $g_m,G$ are gradient-boosted regressors with hyperparameters frozen before evaluation, and
\begin{equation}
\Xpar \;=\; \{a_1,\,b_1,\;\dots,\;a_M,\,b_M\}
\label{eq:xpar}
\end{equation}
collects the parent factors---the raw axes from which each field is constructed. Notation $g(\,\cdot\,;\,e)$ means ``fit to target $e$''.

\textbf{Quotient interpretation of aggregation.} Let the relaxed product space and the zero-sum redistribution subspace be
\begin{equation}
\mathfrak H_{\mathrm{typed}}
:=\prod_{m=1}^{M}\mathcal L_m,
\qquad
\mathcal K_{\Aop}
:=\ker\Aop
=\left\{\mathbf k=(k_1,\ldots,k_M)\in\mathfrak H_{\mathrm{typed}}:
\sum_{m=1}^{M}k_m=0\right\}.
\label{eq:quotient-kernel}
\end{equation}
Define
\begin{equation}
\mathbf H\sim_{\Aop}\mathbf H'
\quad\Longleftrightarrow\quad
\mathbf H-\mathbf H'\in\mathcal K_{\Aop}
\quad\Longleftrightarrow\quad
\Aop(\mathbf H)=\Aop(\mathbf H'),
\qquad
\mathcal H_{\mathrm{quot}}
:=\mathfrak H_{\mathrm{typed}}/\mathcal K_{\Aop}.
\label{eq:quotient-class}
\end{equation}
Thus aggregation identifies \emph{exactly} those typed configurations that differ only by zero-sum redistribution across fields; for example,
\[
(H_1,H_2,H_3)\sim_{\Aop}(H_1+\delta,H_2-\delta,H_3).
\]

\begin{proposition}[Quotient boundary of aggregation]
\label{prop:quotient-boundary}
The map
\begin{equation}
\overline{\Aop}:\mathcal H_{\mathrm{quot}}\longrightarrow\operatorname{Im}\Aop,
\qquad
\overline{\Aop}\bigl([\mathbf H]\bigr):=\Aop(\mathbf H),
\label{eq:quotient-isomorphism}
\end{equation}
is a linear isomorphism. Moreover, holding shared side information fixed, a candidate downstream rule $\mathcal Q:\mathfrak H_{\mathrm{typed}}\to\mathcal Y$ is a legal post-aggregation operator if and only if
\begin{equation}
\mathcal Q(\mathbf H+\mathbf k)=\mathcal Q(\mathbf H)
\quad\forall\,\mathbf k\in\mathcal K_{\Aop}.
\label{eq:quotient-legality}
\end{equation}
Equivalently, there exists a unique induced map
\begin{equation}
\widetilde{\mathcal Q}:\mathcal H_{\mathrm{quot}}\to\mathcal Y,
\qquad
\mathcal Q=\widetilde{\mathcal Q}\circ\pi,
\label{eq:quotient-factorization}
\end{equation}
where $\pi(\mathbf H)=[\mathbf H]$; through Eq.~(\ref{eq:quotient-isomorphism}), this is the same as a unique map on $\operatorname{Im}\Aop$ composed with $\Aop$.
\end{proposition}

\emph{Proof.} The fibers of the linear mean $\Aop$ are precisely the cosets of $\ker\Aop$. The first isomorphism theorem gives Eq.~(\ref{eq:quotient-isomorphism}), and constancy on each fiber is necessary and sufficient for the stated unique factorization. \hfill$\square$

\textbf{Rumination admissibility.} Equations~(\ref{eq:quotient-legality})--(\ref{eq:quotient-factorization}) provide the legality criterion used by both rumination loci below. Rumination-B constructs correction directions only from quotient-surviving shared objects, whereas Rumination-H reads its dynamic discrepancy through quotient-legal maps before refluxing the resulting shared correction back to typed relaxation. Quotient legality is necessary but not sufficient for rumination: contrast or defect structure, $C_0$ centering, residual provenance, and---for Rumination-H---diagonal reflux impose additional mechanism-specific conditions in \S\S\ref{sec:rumination}--\ref{sec:rumination-h}. In short, the quotient does not define rumination; it defines the space in which rumination is allowed to exist.

The post-aggregation backbone is therefore not the original tuple: it is the shared-coordinate realization of its quotient class,
\begin{equation}
\mathbf H\longmapsto[\mathbf H]\cong H=\Aop(\mathbf H).
\label{eq:quotient-lineage}
\end{equation}
The crossroads is the observable realization of a quotient state. This quotient erases the field index, but does not by itself erase the construction provenance carried by the interface: retaining $\bar F$ beside $B=\bar F+\widehat\rho$ keeps $B-\bar F=\widehat\rho$ residual-born. A legal downstream operator may use those surviving shared objects, but it may not recover a coordinate inside $\mathcal K_{\Aop}$.

\textbf{Residual telescoping: what each stage actually fits.} The routing has an exact algebra. Per-field fresh residuals average to the aggregate fresh residual, $\tfrac1M\sum_m (r-F_m)=r-\bar F$ with $\bar F:=\tfrac1M\sum_m F_m$, so---writing $\widehat{(\cdot)}$ for the fitted estimate and defining the aggregate local correction $\widehat{(r-\bar F)}:=\tfrac1M\sum_m\widehat{(r-F_m)}$---aggregation collapses the corrected fields into representation-plus-residual, $\BPQ=\bar F+\widehat{(r-\bar F)}$. Feeding this to the closure, $\bar F$ and the local estimate \emph{cancel} inside the shared target, leaving a residual of the residual:
\begin{equation}
\widehat{S}=\BPQ+G\big(\Xpar;\,r-\BPQ\big)=\bar F+\widehat{\rho}+\widehat{(\rho-\widehat{\rho})},
\qquad \rho:=r-\bar F,\ \ \widehat{\rho}:=\widehat{(r-\bar F)}.
\label{eq:telescope}
\end{equation}
The shared learner never re-sees the fields or their corrections: its target $r-\BPQ=\rho-\widehat\rho$ is exactly what the local stage failed to remove. The fitted predictor and its terminal error can therefore be written as stage-owned additive terms,
\begin{equation}
r=\underbrace{\bar F}_{\text{representation}}+\underbrace{\widehat{\rho}}_{\text{local closure}}+\underbrace{\widehat{(\rho-\widehat{\rho})}}_{\text{shared closure}}+\underbrace{e_2}_{\text{final residual}},
\qquad e_2=(\rho-\widehat\rho)-\widehat{(\rho-\widehat\rho)},
\label{eq:telescope2}
\end{equation}
a twice-applied ``leftover'' operator on $\rho$: capacity is \emph{partitioned} across stages, not shared. This is stagewise boosting \citep{friedman2001greedy} with typed intermediate targets; post-hoc probes find no recoverable out-of-sample structure in $e_2$ within the tested local and shared hypothesis classes (App.~\ref{app:ordering}).

\textbf{Control-variate aggregation and bias correction.} Equation~(\ref{eq:mean}) is not merely an ensemble average. Because $H_m=F_m+\widehat{(r-F_m)}$, it has the exact interface
\begin{equation}
\BPQ=\frac1M\sum_{m=1}^{M}H_m
=\bar F+\widehat{(r-\bar F)}
=\bar F+\widehat\rho,
\label{eq:control-variate}
\end{equation}
where $\bar F$ is a stable low-dimensional anchor and $\widehat\rho$ is a regression adjustment. This is the learned analogue of a control variate \citep{nelson1987control}: the anchor carries the predictable level and the local residual fit removes variation left around it. In the population idealization $\widehat\rho=\Pi_{\mathcal L}\rho$, the $L^2$ projection of $\rho$ onto the aggregate local correction class, the Pythagorean identity gives
\begin{equation}
\mathbb E\!\left[(r-\BPQ)^2\right]
=\mathbb E\!\left[(\rho-\Pi_{\mathcal L}\rho)^2\right]
=\mathbb E[\rho^2]-\mathbb E\!\left[(\Pi_{\mathcal L}\rho)^2\right]
\leq \mathbb E[\rho^2].
\label{eq:variance-reduction}
\end{equation}
Thus the population operator weakly reduces residual second moment relative to $\bar F$; finite regularized GBDTs approximate rather than guarantee this projection identity.

\textbf{Neyman-style orthogonalization.} For a backbone $B$, define $c_B(x):=\mathbb E[r-B\mid\Xpar=x]$ and $S_B:=B+c_B(\Xpar)$, so $\mathbb E[S_B]=\mathbb E[r]$. Along a perturbation $B_t=B+t h$ with the closer refit to the corresponding fresh residual,
\begin{equation}
\left.\frac{\mathrm d}{\mathrm dt}\,
\mathbb E\!\left[B_t+c_{B_t}(\Xpar)\right]\right|_{t=0}
=\mathbb E[h]-\mathbb E\!\left[\mathbb E(h\mid\Xpar)\right]=0.
\label{eq:neyman-path}
\end{equation}
The backbone perturbation is cancelled by the induced change in the residual target. This is Neyman orthogonality only along the \emph{coupled refit path} \citep{chernozhukov2018double}; it does not cover independent perturbations of both learned functions or imply semiparametric efficiency.

\subsection{Rumination-B: finite correction at the crossroads}
\label{sec:rumination}

By Proposition~\ref{prop:quotient-boundary}, any post-aggregation rumination direction must be well defined on quotient classes.

\begin{figure}[t]
\centering
\includegraphics[width=\linewidth]{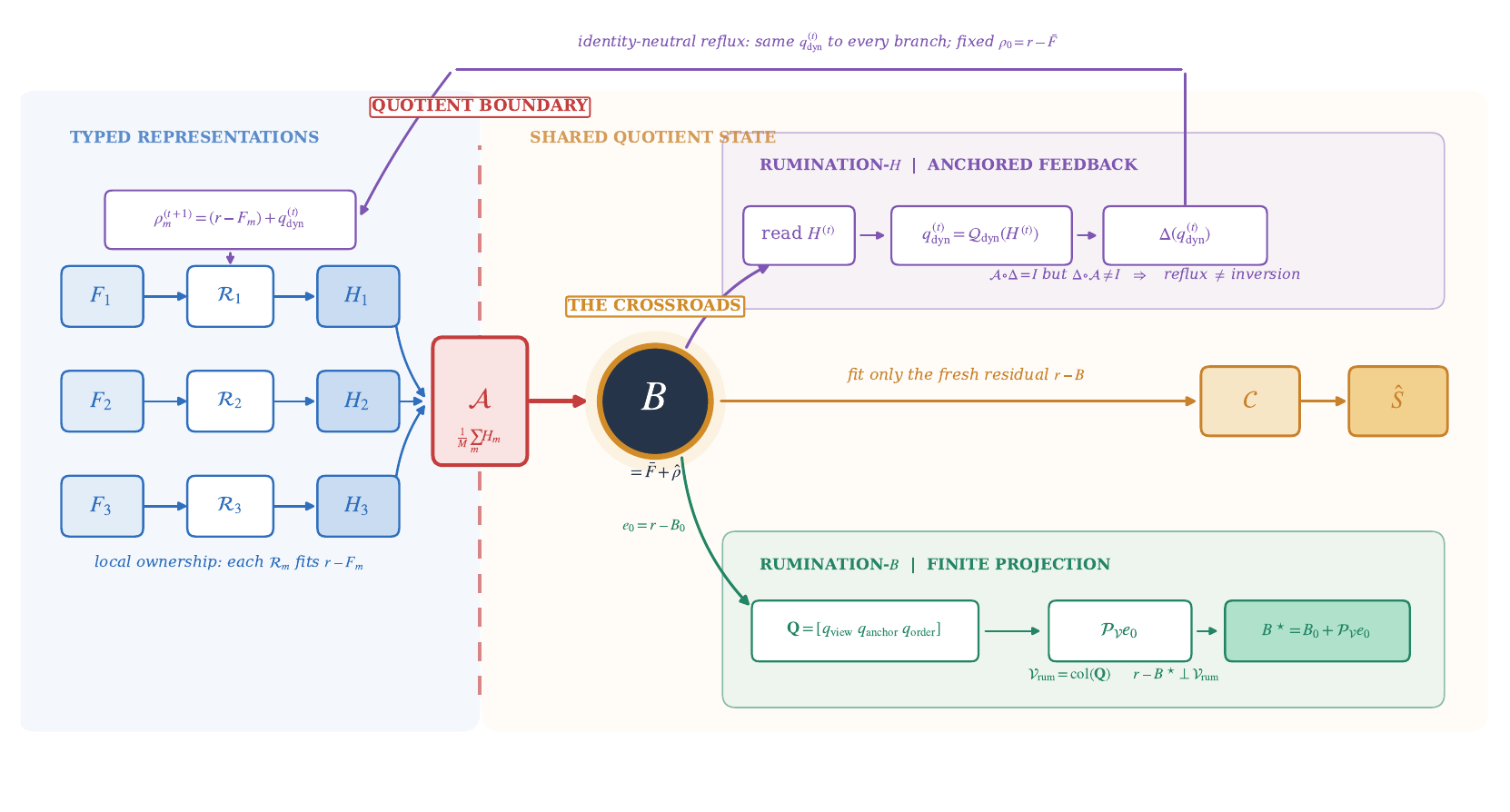}
\caption{\textbf{$B$ is the crossroads of the residual algebra.} \emph{Left:} each typed field fits its own residual before aggregation $\mathcal{A}$ crosses the quotient boundary, removes $\ker\mathcal{A}$, and exposes only the shared backbone $B=\bar F+\widehat\rho$. Closure $\mathcal{C}$ then fits only $r-B$. \emph{Upper lane:} Rumination-$H$ reads $H^{(t)}$, diagnoses $q_{\mathrm{dyn}}^{(t)}$, and uses the diagonal section $\Delta$ to return the same shared correction to every branch at the fixed anchor $\rho_0=r-\bar F$ before typed relaxation is rerun. \emph{Lower lane:} Rumination-$B$ assembles the quotient-legal frame $\mathbf Q$, projects $e_0$ onto $\mathcal V_{\mathrm{rum}}=\operatorname{col}(\mathbf Q)$, and updates $B_0$ to $B^\star$. Since $\mathcal A\!\circ\!\Delta=I$ but $\Delta\!\circ\!\mathcal A\neq I$, reflux does not reconstruct erased identity.}
\label{fig:crossroads}
\end{figure}

The residual algebra admits a second, analytical allocation layer. Let
\[
F:=\bar F,\qquad B_0:=\BPQ=F+\widehat\rho,\qquad e_0:=r-B_0,
\]
where $F$ is the aggregate anchor and $B_0$ the corrected backbone, hence the shared-coordinate realization of a quotient class. The backbone is a \emph{crossroads}, not an archive: typed routes end there, shared routes begin there, and no route may recover the field identity erased by $\Aop$ (Fig.~\ref{fig:crossroads}). FPRC first quotients out pre-aggregation zero-sum redistribution; Rumination-B then constructs new contrasts among the surviving shared objects. A post-aggregation map may therefore read only $(F,B_0,X_{\mathrm{sh}})$, with $X_{\mathrm{sh}}=\Xpar$, and each direction below is legal because it is well-defined on quotient classes. Let $C_0:=I-\Pi_0$ remove the constant mode; every learned map below is a frozen past-data or out-of-fold target-to-prediction operator. The two zero-sum notions are distinct: $\mathcal K_{\Aop}$ is zero-sum across field components and is annihilated by aggregation, whereas $C_0$ centers a surviving shared object across observations.

\textbf{Rumination primitive.} The minimal primitive of a rumination direction is a reconstruction discrepancy. Let $Z$ denote a quotient-surviving shared object and let $\mathcal T_a,\mathcal T_b$ be two quotient-legal routes that read that same object. Their common grammar is the centered contrast
\[
q=C_0\!\left[\mathcal T_a(Z)-\mathcal T_b(Z)\right].
\]
The elementary case $C_0(Z-\widehat Z)=C_0(I-\mathcal T)Z$ takes one route to be the identity and $\widehat Z=\mathcal T(Z)$; replacing the identity by a second route compares two reconstructions, while composing the routes in opposite orders yields the path-order defect $C_0[(\mathcal T_a\!\circ\!\mathcal T_b-\mathcal T_b\!\circ\!\mathcal T_a)Z]$. Thus the hierarchy is identity defect $\subset$ route contrast $\subset$ path-order contrast. The quotient determines what can be seen; rumination compares its different legal reconstructions. The three directions below are instances of this common grammar.

Three contrasts expose what remains negotiable at the crossroads. First, the typed route and a shared route give two views of the same anchor residual $\rho:=r-F$:
\begin{equation}
q_{\mathrm{view}}
:=C_0\!\left[(B_0-F)-\mathcal T_\rho(r-F)\right],
\qquad
\mathcal T_\rho(z):=G_\rho(X_{\mathrm{sh}};z).
\label{eq:rumination-view}
\end{equation}
This isolates disagreement between local ownership and identity-erased reconstruction. Second, the former reflective direction becomes one member of the frame:
\begin{equation}
q_{\mathrm{anchor}}
:=C_0\!\left[F-\mathcal T_F(F)\right],
\qquad
\mathcal T_F(z):=G_F(X_{\mathrm{sh}};z).
\label{eq:rumination-anchor}
\end{equation}
It records what shared reconstruction displaces from the anchor. Third, let $\mathcal T_a,\mathcal T_b$ be two frozen shared reconstruction operators on the same post-aggregation state. Their path-order defect is
\begin{equation}
q_{\mathrm{order}}
:=C_0\!\left[
\mathcal T_a\!\bigl(\mathcal T_b(B_0)\bigr)
-\mathcal T_b\!\bigl(\mathcal T_a(B_0)\bigr)
\right].
\label{eq:rumination-order}
\end{equation}
It vanishes when the routes commute on the tested state. For nonlinear refitted learners it is a composition-order defect---a commutator only in the linear case.

\begin{figure}[t]
\centering
\includegraphics[width=\linewidth]{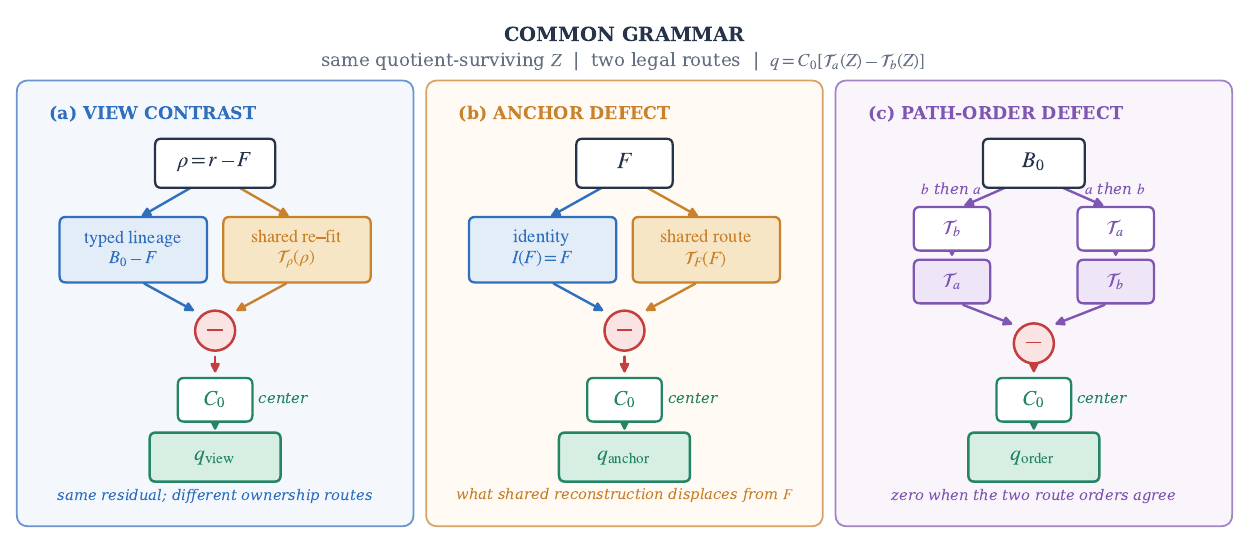}
\caption{\textbf{Three Rumination-$B$ directions are centered discrepancies between two quotient-legal readings of the same surviving object.} \textbf{(a)}~$q_{\mathrm{view}}$ compares the typed-lineage residual estimate $B_0-F$ with a shared reconstruction $\mathcal T_\rho(\rho)$ of the same $\rho=r-F$. \textbf{(b)}~$q_{\mathrm{anchor}}$ compares the identity reading $F$ with its shared reconstruction $\mathcal T_F(F)$, measuring anchor displacement. \textbf{(c)}~$q_{\mathrm{order}}$ compares the two route orders $\mathcal T_a\!\circ\!\mathcal T_b$ and $\mathcal T_b\!\circ\!\mathcal T_a$ on $B_0$, and vanishes when they agree on the tested state. In every panel, the red node takes the ordered difference, $C_0$ removes the constant mode across observations, and the green output becomes one column of $\mathbf Q$.}
\label{fig:rumination-directions}
\end{figure}

Collect the directions as the synthesis operator
\begin{equation}
\mathbf Q:=[q_{\mathrm{view}},q_{\mathrm{anchor}},q_{\mathrm{order}}],
\qquad
\mathcal V_{\mathrm{rum}}:=\operatorname{col}(\mathbf Q),
\qquad
B_{\boldsymbol\lambda}:=B_0+\mathbf Q\boldsymbol\lambda .
\label{eq:rumination-frame}
\end{equation}
The three columns are not three features to tune independently; they are a basis for the legal innovation space left at $B_0$. Figure~\ref{fig:rumination-directions} makes their common contrast grammar and distinct route semantics explicit.

\begin{proposition}[Optimal crossroads rumination]
\label{prop:rumination}
Take inner products over the point-in-time fitting sample and let $(\cdot)^+$ denote the Moore--Penrose inverse. The minimum-norm analytical coefficient and induced backbone are
\begin{equation}
\boldsymbol\lambda^\star
=(\mathbf Q^\ast\mathbf Q)^+\mathbf Q^\ast e_0,
\qquad
B^\star
=B_0+\mathbf Q\boldsymbol\lambda^\star
=B_0+\mathcal P_{\mathcal V_{\mathrm{rum}}}e_0,
\label{eq:rumination-projection}
\end{equation}
where
$\mathcal P_{\mathcal V_{\mathrm{rum}}}
:=\mathbf Q(\mathbf Q^\ast\mathbf Q)^+\mathbf Q^\ast$.
Then $B^\star$ is the unique element of $B_0+\mathcal V_{\mathrm{rum}}$ minimizing $\lVert r-B\rVert^2$, and
\begin{equation}
r-B^\star
=(I-\mathcal P_{\mathcal V_{\mathrm{rum}}})e_0
\perp\mathcal V_{\mathrm{rum}}.
\label{eq:rumination-ownership}
\end{equation}
Coefficients are unique when the directions are independent; the projected increment and $B^\star$ remain unique otherwise. If $\mathcal V_{\mathrm{rum}}=\{0\}$, the construction recovers $B_0$.
\end{proposition}

\emph{Proof.} The normal equations for $\min_{\boldsymbol\lambda}\lVert e_0-\mathbf Q\boldsymbol\lambda\rVert^2$ are $\mathbf Q^\ast\mathbf Q\boldsymbol\lambda=\mathbf Q^\ast e_0$; their Moore--Penrose solution is the orthogonal projection onto $\operatorname{col}(\mathbf Q)$. \hfill$\square$

Rumination owns the fresh residual's projection onto the three legal directions; terminal closure owns only the complement. A parents-only closer preserves the boundary,
\begin{equation}
\widehat r_{\mathrm{rum}}
=B^\star+G_{\mathrm{close}}\bigl(X_{\mathrm{sh}};r-B^\star\bigr),
\qquad X_{\mathrm{sh}}=\Xpar.
\label{eq:rumination-predictor}
\end{equation}
Its parameter-free mechanism strength is
\begin{equation}
\eta_{\mathrm{rum}}
:=\frac{\lVert\mathcal P_{\mathcal V_{\mathrm{rum}}}e_0\rVert^2}
{\lVert e_0\rVert^2}
=1-\frac{\lVert r-B^\star\rVert^2}{\lVert e_0\rVert^2}.
\label{eq:rumination-efficiency}
\end{equation}
All inner reconstructions, nested compositions, directions, and projection coefficients must be generated on past or strictly out-of-fold observations and frozen before prediction; estimating any of them from evaluation labels would be leakage. This is a theoretical extension: every reported return in this paper evaluates base FPRC-PQ, recovered by the zero-increment member $B^\star=B_0$.

\subsection{Rumination-H: anchored reflux through typed relaxation}
\label{sec:rumination-h}

The same quotient law governs feedback: diagnosis is performed on the quotient state, while reflux uses the symmetric section in Eq.~(\ref{eq:rumination-h-section}) and does not invert the quotient.

Rumination-B writes a finite correction at $B_0$. Rumination-H never adds $q$ directly to $H$. It reads a discrepancy from the current aggregate state and returns that discrepancy to the \emph{fixed} residual problem from which the typed branches are relaxed:
\[
\boxed{\qquad \text{read from }H^{(t)},\qquad \text{write to }\rho_0=r-\bar F.\qquad}
\]
Let
\begin{equation}
\rho_0:=r-\bar F,\qquad
H_m^{(0)}:=F_m+g_m^{(0)}\bigl(X_m^{\mathrm{local}};r-F_m\bigr),\qquad
H^{(0)}:=\Aop\bigl(H_1^{(0)},\ldots,H_M^{(0)}\bigr),
\label{eq:rumination-h-init}
\end{equation}
and define the aggregate residual estimate $\rho_H^{(t)}:=H^{(t)}-\bar F$. Thus $\rho_H^{(t)}$ is a state read from the current backbone, whereas $\rho_0$ is the residual anchor to which every update is written.

Only residual-born, quotient-compatible directions may participate in the dynamic loop. Let $\mathcal T_L,\mathcal T_S:\mathcal H_{\mathrm{quot}}\to\mathcal H_{\mathrm{quot}}$ be two frozen residual reconstruction routes on the same identity-erased state: $\mathcal T_L$ is the quotient image of the typed route and $\mathcal T_S$ is a shared route. Two dynamic diagnostics are
\begin{align}
q_{\mathrm{view}}^{(t)}
&:=C_0\bigl(\mathcal T_L-\mathcal T_S\bigr)\rho_H^{(t)},
\label{eq:rumination-h-view}\\
q_{\mathrm{order}}^{(t)}
&:=C_0\bigl(\mathcal T_L\!\circ\!\mathcal T_S
-\mathcal T_S\!\circ\!\mathcal T_L\bigr)\rho_H^{(t)}.
\label{eq:rumination-h-order}
\end{align}
The second quantity is a residual-side path-order defect, and a commutator only when the routes are linear. Their admissible combination is
\begin{equation}
q_{\mathrm{dyn}}^{(t)}
:=\mathcal Q_{\mathrm{dyn}}\bigl(H^{(t)}\bigr)
=\lambda_1^{(t)}q_{\mathrm{view}}^{(t)}
+\lambda_4^{(t)}q_{\mathrm{order}}^{(t)}.
\label{eq:rumination-h-dynamic}
\end{equation}
The coefficients are fixed or produced by one prespecified past-data/out-of-fold rule, which is included in $\mathcal Q_{\mathrm{dyn}}$; they are never selected from evaluation returns. In contrast, the anchor-born direction
\begin{equation}
q_{\mathrm{static}}:=C_0\bigl[\bar F-\mathcal T_F(\bar F)\bigr]
\label{eq:rumination-h-static}
\end{equation}
is excluded from $\mathcal Q_{\mathrm{dyn}}$. This is a type restriction, not a heuristic split: the quotient removes representation identity, not the semantic provenance retained by the shared interface. Here $\rho_H^{(t)}=H^{(t)}-\bar F$ is residual-born, whereas $q_{\mathrm{static}}$ is anchor-born.

The decisive operation is reflux, not accumulation:
\begin{equation}
\widetilde\rho^{(t)}=\rho_0+q_{\mathrm{dyn}}^{(t)},
\qquad
\rho_m^{(t+1)}=r-F_m+q_{\mathrm{dyn}}^{(t)}
=\widetilde\rho^{(t)}+(\bar F-F_m).
\label{eq:rumination-h-reflux}
\end{equation}
On the common shared correction space, define the symmetric diagonal section
\begin{equation}
\Delta(q):=(q,\ldots,q),
\qquad
\Aop\circ\Delta=I_{\operatorname{Im}\Aop},
\qquad
\Delta\circ\Aop\neq I_{\mathfrak H_{\mathrm{typed}}},
\qquad
\mathbf H-\Delta\Aop(\mathbf H)\in\mathcal K_{\Aop}.
\label{eq:rumination-h-section}
\end{equation}
There is no $1/M$ factor: because $\Aop$ is a mean, the same shared correction must enter every branch at the same amplitude. Equation~(\ref{eq:rumination-h-section}) makes the asymmetry exact. The section selects an identity-neutral representative of the quotient-level innovation, but cannot reconstruct the discarded fieldwise component. Reflux is possible; inversion is not. Each branch then interprets the shared diagnosis in its own coordinates,
\begin{align}
H_m^{(t+1)}
&=F_m+g_m^{(t+1)}\!\left(
X_m^{\mathrm{local}};r-F_m+q_{\mathrm{dyn}}^{(t)}
\right),
\label{eq:rumination-h-branch}\\
H^{(t+1)}
&=\frac1M\sum_{m=1}^M H_m^{(t+1)}
=:\Phi_{\rho_0}\bigl(H^{(t)}\bigr).
\label{eq:rumination-h-update}
\end{align}
Equivalently, if $\mathcal R_{\rho_0}$ denotes the prescribed fit-and-relax operation in Eq.~(\ref{eq:rumination-h-branch}), then
\begin{equation}
\Phi_{\rho_0}
:=\Aop\circ\mathcal R_{\rho_0}\circ\mathcal Q_{\mathrm{dyn}},
\qquad
\mathcal R_{\rho_0}(q)
:=\left(F_m+g_m^+\bigl(X_m^{\mathrm{local}};r-F_m+q\bigr)\right)_{m=1}^M .
\label{eq:rumination-h-composite}
\end{equation}
Equations~(\ref{eq:rumination-h-reflux})--(\ref{eq:rumination-h-composite}), rather than the ambiguous shorthand $\mathcal R[\rho_0+\mathcal Q(H)]$, are the typed definition: the branch-specific offset $\bar F-F_m$ remains intact. In particular, Rumination-H is neither target rewriting, $r^{(t+1)}=r^{(t)}+q^{(t)}$, nor state accumulation, $H^{(t+1)}=H^{(t)}+q^{(t)}$. Its read/write asymmetry is
\begin{equation}
\boxed{\quad H^{(t)}\ \xrightarrow{\ \mathcal Q_{\mathrm{dyn}}\ }\ q_{\mathrm{dyn}}^{(t)}
\quad\text{but}\quad
q_{\mathrm{dyn}}^{(t)}\ \longmapsto\ (r-F_m)+q_{\mathrm{dyn}}^{(t)}.\quad}
\label{eq:rumination-h-asymmetry}
\end{equation}
Thus the structural sequence is \emph{shared diagnosis $\to$ identity-neutral reflux $\to$ typed reinterpretation}.

The iteration stops when $\lVert H^{(t+1)}-H^{(t)}\rVert\leq\varepsilon_H$, optionally together with $\lVert q_{\mathrm{dyn}}^{(t+1)}-q_{\mathrm{dyn}}^{(t)}\rVert\leq\varepsilon_q$. Vanishing correction is not required. A rumination-consistent state satisfies
\begin{equation}
H^\star=\Phi_{\rho_0}(H^\star)
=\Aop\!\left(
\left(F_m+g_m^\star\bigl(X_m^{\mathrm{local}};r-F_m+
\mathcal Q_{\mathrm{dyn}}(H^\star)\bigr)\right)_{m=1}^M
\right),
\label{eq:rumination-h-fixed}
\end{equation}
which may hold with $q_{\mathrm{dyn}}^\star\neq0$: a discrepancy diagnosed from $H^\star$, returned to $\rho_0$, reproduces the same aggregate state after typed relaxation.

\begin{proposition}[Conditional convergence of Rumination-H]
\label{prop:rumination-h}
Suppose $\mathcal Q_{\mathrm{dyn}}$ is $L_Q$-Lipschitz and the induced anchored relaxation-aggregation map $q\mapsto\Aop\mathcal R_{\rho_0}(q)$ is $L_R$-Lipschitz in its shared correction. Then
\begin{equation}
\bigl\lVert\Phi_{\rho_0}(H)-\Phi_{\rho_0}(H')\bigr\rVert
\leq L_RL_Q\lVert H-H'\rVert.
\label{eq:rumination-h-lipschitz}
\end{equation}
If $L_RL_Q<1$, $\Phi_{\rho_0}$ has a unique fixed point and
\begin{equation}
\lVert H^{(t)}-H^\star\rVert
\leq(L_RL_Q)^t\lVert H^{(0)}-H^\star\rVert.
\label{eq:rumination-h-convergence}
\end{equation}
\end{proposition}

\emph{Proof.} Composition gives the Lipschitz bound; the fixed-point and geometric-convergence claims follow from the contraction mapping theorem. \hfill$\square$

This proposition is conditional: finite refitted GBDTs are not asserted to satisfy a contraction. After convergence, static rumination is applied once and only once,
\begin{equation}
B_H^\star:=H^\star+\lambda_s q_{\mathrm{static}},
\qquad
\widehat S_H:=B_H^\star+G_{\mathrm{close}}\bigl(\Xpar;r-B_H^\star\bigr),
\label{eq:rumination-h-close}
\end{equation}
with $\lambda_s$ frozen from past or out-of-fold data. The full chain is therefore
\begin{equation}
\Fvec\xrightarrow{\Rop}\mathbf H^{(0)}\xrightarrow{\Aop}H^{(0)}
\xrightarrow{\text{anchored dynamic reflux}}\cdots\xrightarrow{}H^\star
\xrightarrow{\text{static rumination}}B_H^\star\xrightarrow{\Cop}\widehat S_H.
\label{eq:rumination-h-chain}
\end{equation}
Rumination-B is an algebraic correction at the post-aggregation crossroads; Rumination-H is an anchored nonlinear self-correction that reads from $H$ but writes to $r-\bar F$.

\subsection{Operator principles and scope}

Four algebraic principles are encoded:

\begin{figure}[H]
\centering
\includegraphics[width=\linewidth]{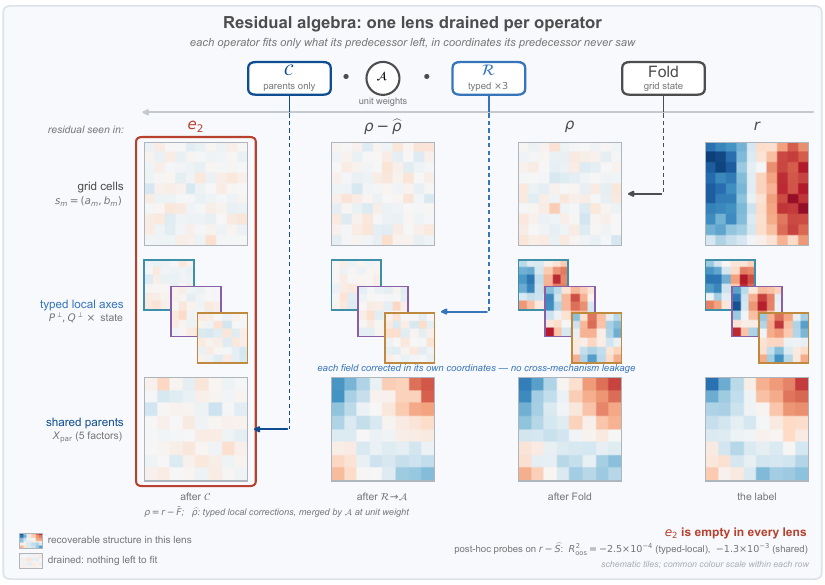}
\caption{\textbf{Residual structure is routed through typed operators.} 
The residual is sequentially consumed by three coordinate systems: field grids $s_m$ (top), typed local axes (middle), and shared parents $\Xpar$ (bottom), following Eq.~\ref{eq:telescope2}. 
\textbf{Fold} removes grid-level structure; $\Rop$ corrects each field in its own coordinates; $\Aop$ aggregates the corrected fields; and $\Cop$ operates only on the remaining aggregate residual. 
The staircase is schematic rather than an orthogonal decomposition. The terminal residual $e_2$ contains no detectable out-of-sample signal within tested local and shared probe classes ($R^2_{\mathrm{oos}}\le0$; App.~\ref{app:ordering}).}
\label{fig:fprc}
\end{figure}

\textbf{(D1) Typed local correction.} Each $g_m$ sees only its own field's residual, expressed in that field's coordinates $(a_m,b_m)$ plus the fixed auxiliary interface. Corrections therefore cannot leak across mechanisms: each error is repaired in the coordinate system of the field that produced it.

\textbf{(D2) Quotient-controlled identity erasure.} Fields enter (\ref{eq:local})--(\ref{eq:mean}) as unit-coefficient offsets; no gate or learned blend re-weights them. Their identities remain intact through relaxation and are erased exactly once: the mean kills $\ker\Aop$, and every legal downstream operator must factor through $\mathcal H_{\mathrm{quot}}$. Aggregate drift therefore cannot silently reassign responsibility between mechanisms.

\textbf{(D3) Fresh-residual closure.} The shared stage fits $r-\BPQ$---the error remaining \emph{after} local semantics are exhausted---from parent factors only. It neither re-reads local states nor the aggregate's own rank (both were tested and rejected; \S\ref{sec:results-audit}).

\textbf{(D4) Staged information compression.} Semantically local information (grid coordinates and fixed auxiliaries) is consumed at the local stage; only generic, field-agnostic coordinates remain for closure. Capacity is placed where the corresponding error population lives, not where it is easiest to add.

\textbf{Scope of the projection view.} The Hilbert-space argument belongs to Fold's population interface, not to the fitted ensemble as an exact orthogonal decomposition. Empirical fields and both correctors are finite, regularized learners; the field subspaces overlap; and $\Aop$ is not a projection onto their sum. Equations~(\ref{eq:local})--(\ref{eq:closure}) should therefore be read as an ordered, finite-capacity routing hypothesis whose targets, boundaries, and order are tested empirically in \S\ref{sec:results-audit}.

FPRC-PQ is deliberately \emph{not} defended by a larger capacity budget. In the matched control, one direct GBDT receives exactly the sum of the three local learners' and the shared closer's trees in each deployment year; it still trails the typed composition (App.~\ref{app:model-controls}). The architecture was not designed by intuition alone: five preregistered audits, each freezing a question and a decision rule before fitting, selected (\ref{eq:local})--(\ref{eq:closure}) over its neighbors (\S\ref{sec:results-audit}, Table~\ref{tab:audit}).

\section{Experimental setup}
\label{sec:setup}

\textbf{Data and protocol.} We evaluate on the Chinese A-share market \citep{leippold2022machine}: 844 trading days from 2023-01 to 2026-07, median 4{,}380 stocks per day, for a strict common book of 3{,}666{,}920 stock-day rows (a wider ``available'' book of 3{,}690{,}036 rows serves as a coverage-sensitivity check; conclusions must agree on both). The label is the 5-day forward return. Deployment is annual point-in-time: models for year $Y$ are fit on data observable before $Y$ with a purge gap covering label observability \citep{lopez2018advances}. \emph{Every} selection---field definitions, auxiliary coordinates, learner settings, and architecture---was frozen on data through 2022-12-31; no 2023--2026 label influenced any choice.

\textbf{Learners and execution.} $g_m$ and $G$ are XGBoost regressors with frozen settings (depth 5, learning rate 0.035, $\le$600 trees with early stopping on the preceding calendar year, min-child-weight 2000, $\lambda{=}10$). Signals are executed long-only (dropping the bottom quintile), with 21-day phased holdings and blended 10/15\,bp transaction costs; we report cumulative net-of-cost return over the 42-month window, annualized Sharpe on daily net returns, and validation ICIR.

\textbf{Status of rumination.} The headline experiments use the base backbone $\BPQ$ (zero rumination increment); no reported return is attributed to Rumination-B or Rumination-H in \S\S\ref{sec:rumination}--\ref{sec:rumination-h}. A future deployment must generate every direction, coefficient, iteration rule, and stopping decision strictly from past or out-of-fold data and freeze the entire mechanism before evaluation.

\textbf{Inference.} All comparisons are \emph{paired} on identical books and identical dates, with 21-day block bootstrap (10{,}000 draws) \citep{kunsch1989jackknife,politis1994stationary}; we report observed deltas in percentage points (pp), $\Pr[\Delta>0]$, and 95\% intervals. Because multiple preregistered variants exist, no method is promoted merely for ranking first \citep{harvey2016cross,bailey2017probability}; verdicts follow the registered contrasts. Reproduction audits require every frozen signal to be re-derivable bit-exactly (max difference $0$; App.~\ref{app:audit-ledger}).

\section{Results}
\label{sec:results}

\subsection{The architecture ladder: placement beats pooling and beats size}
\label{sec:results-ladder}

\begin{figure}[t]
\centering
\includegraphics[width=\linewidth]{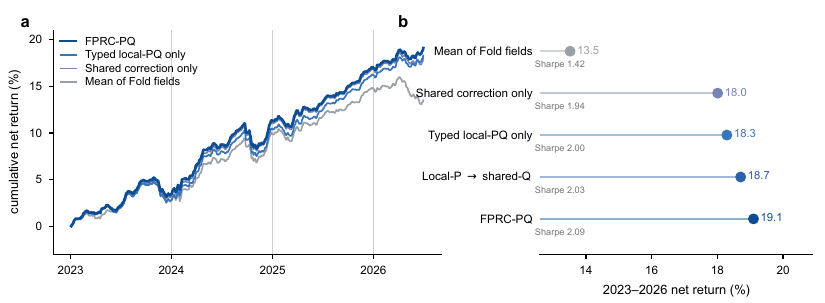}
\caption{\textbf{(a)}~Cumulative net-of-cost return, strict book. Curves separate where the residual algebra predicts: typed local correction over pooling, then fresh closure over either stage alone. \textbf{(b)}~The same architecture ladder with annualized Sharpe.}
\label{fig:performance}
\end{figure}

Table~\ref{tab:ladder} and Fig.~\ref{fig:performance} trace the ladder from raw representation to full architecture on the strict book. Three comparisons isolate the design principles.

\begin{table}[t]
\caption{\textbf{Architecture ladder}, strict book (3.67M rows, 2023--2026). Net is cumulative net-of-cost return; $\Delta$ is versus the pooled-field baseline (Mean of three Fold fields). All rows share the same data, 10/15\,bp cost model, and frozen learner hyperparameters. The top block gives two \emph{same-features} baselines: a single gradient-boosted learner fit to the full pooled feature set $X_{\mathrm{full}}$ (all factors and their derived inputs), either directly on the label ($G(X_{\mathrm{full}};r)$) or on the pooled-field residual ($\bar F+G(X_{\mathrm{full}};r-\bar F)$)---the same inputs, without Fold state or typed composition. The \emph{Operator} column of the remaining rows composes Eqs.~(\ref{eq:local})--(\ref{eq:closure}): $\Aop$ is equal-mean aggregation, $\mathcal{R}_{\bullet}$ a typed local relaxation seated on axes $\bullet$ ($\mathrm{P}$ or $\mathrm{PQ}$), $\Cop$ a parents-only fresh-residual closure, and $\mathcal{C}_{Q}$ a closure that incorrectly rereads $Q$.}
\label{tab:ladder}
\begin{center}
\small
\setlength{\tabcolsep}{3pt}
\begin{tabular}{llcccccc}
\toprule
Model & Operator & ICIR & Sharpe & Gross\% & Cost\% & Net\% & $\Delta$ (pp) \\
\midrule
Direct GBDT, same features & $G(X_{\mathrm{full}};\,r)$ & 4.74 & 1.67 & 18.64 & 2.72 & 15.93 & $+2.41$ \\
Residual GBDT, same features & $\bar F + G(X_{\mathrm{full}};\,r-\bar F)$ & 4.94 & 1.79 & 19.68 & 2.70 & 16.98 & $+3.46$ \\
\midrule
Mean of three Fold fields & $\Aop$ & 4.45 & 1.42 & 16.09 & 2.57 & 13.52 & --- \\
Shared correction only (on $\bar F$) & $\Cop\!\circ\!\Aop$ & 4.90 & 1.94 & 20.63 & 2.62 & 18.01 & $+4.49$ \\
Typed local-P only & $\Aop\!\circ\!\mathcal{R}_{\mathrm P}$ & 5.09 & 1.94 & 20.45 & 2.43 & 18.02 & $+4.50$ \\
Typed local-PQ only & $\Aop\!\circ\!\Rop$ & 5.08 & 2.00 & 20.73 & 2.44 & 18.29 & $+4.78$ \\
Local-P $\to$ shared-Q (predecessor) & $\mathcal{C}_{Q}\!\circ\!\Aop\!\circ\!\mathcal{R}_{\mathrm P}$ & 5.03 & 2.03 & 21.30 & 2.59 & 18.71 & $+5.19$ \\
\textbf{FPRC-PQ} & $\Cop\!\circ\!\Aop\!\circ\!\Rop$ & 5.01 & \textbf{2.09} & 21.67 & 2.57 & \textbf{19.10} & $+5.58$ \\
\quad + shared-Q reread & $\mathcal{C}_{Q}\!\circ\!\Aop\!\circ\!\Rop$ & 5.04 & 2.07 & 21.37 & 2.57 & 18.80 & $+5.28$ \\
\bottomrule
\end{tabular}
\end{center}
\end{table}

\emph{Capacity control.} On the same 3{,}666{,}920-row strict book, FPRC-PQ earns 19.098\% net with Sharpe 2.089. A direct learner given exactly the combined annual tree budget of the three local learners and the shared closer reaches only 16.970\% and Sharpe 1.963, trailing by $2.128\pp$ and losing in all four years. Extra direct capacity raises Sharpe relative to the ordinary direct control (1.882) but does not close the structural gap (App.~\ref{app:model-controls}).

\emph{Identity control.} Collapsing the three fields into one concatenated representation and fitting a unified residual gives 16.824\% net, $2.273\pp$ below FPRC-PQ in 4/4 years (95\% CI $[+0.123,+4.453]\pp$ for FPRC minus control). A fresh but identity-free two-stage booster also fails (17.408\%, $-1.690\pp$, 0/4 years). Thus two-stage boosting alone is insufficient: the intermediate target must retain which field produced it, and each correction must be fitted to that field's own residual.

\emph{Interaction control.} An explicit pairwise-only histogram GBDT---a GA$^2$M-class control, not an EBM package benchmark---reaches 16.681\% net and Sharpe 1.939, trailing FPRC-PQ by $2.417\pp$ in 4/4 years. The gain is not explained by total tree capacity, pairwise interaction modeling, or generic two-stage boosting. It appears only when each field retains its identity and receives a correction fitted to its own residual before aggregation.

\subsection{Five frozen audits: why the closure looks the way it does}
\label{sec:results-audit}

Each structural choice in (\ref{eq:local})--(\ref{eq:closure}) was selected by a preregistered audit with a frozen question, decision rule, and paired inference (Table~\ref{tab:audit}; full tables in App.~\ref{app:audit}).

\begin{table}[t]
\caption{\textbf{The operator decision chain.} Five frozen audits; contrasts are paired block-bootstrap deltas in pp on the strict book (95\% CI). Positive favors the selected structure.}
\label{tab:audit}
\begin{center}
\scriptsize
\setlength{\tabcolsep}{3.5pt}
\begin{tabular}{p{0.35cm}p{4.35cm}p{3.2cm}p{4.6cm}}
\toprule
\# & Frozen question & Selected structure & Key contrast \\
\midrule
A1 & Does typed local correction beat pooling and global-only correction? & local identity retained before aggregation & typed local-PQ vs.\ pooled fields: $+4.78$; composed vs.\ best single stage: $+0.81$ \\
A2 & Does the equal-mean interface discard state the closer needs? & plain mean; no disagreement coordinates & best disagreement variant vs.\ mean: $+0.001$ $[-0.17,0.17]$, $\Pr=0.47$ \\
A3 & Should cross-level edges (shared-$P$, dual-$Q$) be reopened? & asymmetric routing; no full closure & full reopening vs.\ FPRC routing: $-0.12$ $[-0.89,0.58]$, $\Pr=0.31$ \\
A4 & Does $Q$ belong locally, shared, or both? & local $Q$; shared slot closed & local-only vs.\ dual-layer $Q$: $+0.30$ $[-0.15,0.73]$, $\Pr=0.91$; interaction $-0.31$ $[-0.35,-0.01]$ \\
A5 & May the closer read the aggregate's own rank $u_B$? & empty shared state slot & self-state vs.\ empty: $-0.45$ $[-0.80,-0.05]$, $\Pr=0.02$ \\
\bottomrule
\end{tabular}
\end{center}
\end{table}

Three findings deserve emphasis. First, the \emph{negative interaction} in A4 ($\Gamma=-0.31\pp$, CI $[-0.35,-0.01]$): once $Q$ is consumed locally, re-exposing it at the shared stage is redundant and mildly destructive---capacity placement is a genuine either/or, not an additive menu (App.~Fig.~\ref{fig:placement}). Second, A5 is a \emph{significant} rejection: letting the closer see the aggregate signal's own cross-sectional rank over-calibrates ($-0.45\pp$, and $-0.38\pp$ versus its permutation control), justifying D3's strict fresh-residual discipline.

Third, A2 is an empirical stress test of the quotient boundary. The discarded component
\begin{equation}
\mathbf K_{\mathrm{disc}}(\mathbf H)
:=\mathbf H-\Delta\Aop(\mathbf H)
\in\ker\Aop
\label{eq:discarded-kernel}
\end{equation}
is probed by orthogonal disagreements between corrected experts. Re-exposing the best tested disagreement coordinate after the mean changes net return by only $+0.001\pp$ (CI $[-0.17,0.17]$, $\Pr[\Delta>0]=0.47$), so the killed field-redistribution component carries no measurable incremental trading information in the tested class. This is empirical support for the chosen quotient, not a universal sufficiency theorem. Relaxing the unit coefficients likewise moves within noise: an annually-refit optimal common scaling gives $-0.01\pp$, a two-coefficient analytic reweighting gives $+0.05\pp$, and doubling the anchor harms performance. Thus the equal mean is both quotient-clean and locally flat in performance (App.~\ref{app:ordering}).

\subsection{Boundary controls: side information cannot repair the wrong algebra}
\label{sec:results-controls}

\begin{figure}[t]
\centering
\includegraphics[width=\linewidth]{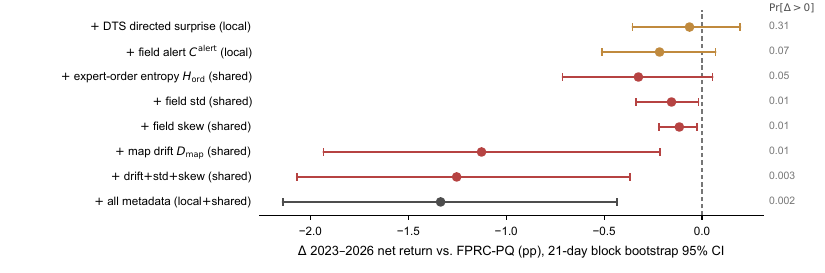}
\caption{\textbf{Eight failed shortcuts.} Paired deltas versus frozen FPRC-PQ with 95\% block-bootstrap intervals, strict book. Amber denotes local channels and red shared channels; the right column gives $\Pr[\Delta>0]$. Error-correlated metadata does not improve the model merely by being appended to a correction stage.}
\label{fig:controls}
\end{figure}

The positive result is not a license to append arbitrary state to a strong backbone. Eight preregistered channels---a local outlier alert $C^{\mathrm{alert}}$, a market-level map-drift scalar $D_{\mathrm{map}}$ (short- vs.\ long-window field discrepancy), committee dispersion (std) and shape (skew), shared order entropy, transition surprise, and their combinations---produce deltas from $-0.06$ to $-1.34\pp$; five are significantly negative (Fig.~\ref{fig:controls}). The drift scalar is instructive: it correlates with committee-level absolute error ($\rho=0.24$) yet \emph{destroys} $1.13\pp$ when the closer conditions on it, collapsing the closer's fitted complexity (mean trees $51\to17$). A covariate can diagnose error and still be harmful when exposed after identity erasure. Extra information is not a substitute for a correctly typed residual target.

\subsection{Robustness and audits}
\label{sec:results-robust}

The architecture conclusions replicate on the wider available book and across deployment years (App.~\ref{app:yearly}), including the purely forward 2026 stub. Frozen-signal reproduction audits pass bit-exactly; native-missing gates, purge gaps, and point-in-time label availability are checked per feature (max reproduction difference $0.0$; App.~\ref{app:audit-ledger}). Wall-clock cost is modest: the full algebra fits in minutes per year on four GPU workers.

\section{Related work}
\label{sec:related}

\textbf{Tabular learning.} GBDTs \citep{friedman2001greedy,chen2016xgboost,ke2017lightgbm,prokhorenkova2018catboost} remain the reference on medium-to-large tabular tasks \citep{grinsztajn2022tree,mcelfresh2023neural,shwartzziv2022tabular}; deep alternatives \citep{arik2021tabnet,gorishniy2021revisiting,hollmann2025tabpfn} narrow the gap partly by learning feature discretizations \citep{gorishniy2022embeddings}. Fold makes the discretization explicit, supervised only by rank structure, and shared across all downstream stages. Interaction-explicit models (GA$^2$M/EBM, NAM) \citep{lou2013accurate,caruana2015intelligible,agarwal2021neural} pursue intelligibility for pairwise effects; Fold differs in making the interaction a \emph{representation} that owns an intermediate residual, not merely a final additive term.

\textbf{Experts, stacking, and residual staging.} FPRC relates to mixtures of experts \citep{jacobs1991adaptive,jordan1994hierarchical,shazeer2017outrageously} and stacking \citep{wolpert1992stacked}, but inverts their premise: experts are defined by fixed interpretable representations rather than learned gates; aggregation is an explicit identity-erasure boundary; and the audit trail (\S\ref{sec:results-audit}) treats \emph{where} residual capacity lives as the experimental object. The relax--aggregate--close composition is closest to stagewise boosting \citep{friedman2001greedy}, but differs by making intermediate targets representation-typed and by forbidding later stages from rereading consumed local state.

\textbf{Financial ML.} Large-scale return prediction with ML is well established \citep{gu2020empirical,kelly2019characteristics,leippold2022machine}, as are warnings on multiple testing and leakage \citep{harvey2016cross,bailey2017probability,lopez2018advances}. Our contribution to this literature is methodological: a frozen-protocol architecture study with preregistered controls, rather than a new predictor hunt. The name ``factor fields'' is also used for neural-field basis decompositions in vision \citep{chen2023factor}; the two lines are unrelated---our fields are discrete conditional-mean tables over tabular rank grids.

\section{Conclusion}
\label{sec:conclusion}

The central result is that representation learning benefits from an explicit residual algebra. Fold turns each interaction into a typed conditional-mean object; FPRC applies a local residual operator while that type is intact, erases identity only at a fixed aggregation boundary, and closes the aggregate with a genuinely fresh target. The boundary is formal: $\mathcal H_{\mathrm{quot}}=\mathfrak H_{\mathrm{typed}}/\ker\Aop\cong\operatorname{Im}\Aop$, so legal shared operators are exactly those that factor through the quotient. This ordering gives an exact telescoping decomposition and a control-variate correction, and it moves the frozen system from 13.52\% to 19.10\% net (Sharpe 1.42${\to}$2.09). A2 finds no measurable gain from returning corrected-expert disagreement after the mean, empirically supporting---without universally proving---the chosen quotient. Capacity-matched, unified-residual, identity-free, pairwise-only, learned-aggregation, and side-information controls further delimit the explanation: the gain does not follow from more trees, generic staging, or feature abundance. Rumination-B constructs quotient-legal view, anchor, and path-order contrasts only after the old field-redistribution freedom has been killed. Rumination-H reads residual-born diagnosis from $H$ and uses the symmetric section $\Delta$ to reflux it to the immutable anchor $r-\bar F$; because $\Aop\Delta=I$ but $\Delta\Aop\neq I$, reflux is possible without pretending to invert identity erasure. Anchor-born rumination enters only after convergence. Neither extension contributes to the reported empirical headline, and both remain to be tested out of sample. Evidence still comes from one market and one asset class; the fields are two-dimensional; the contraction statement is conditional; and the orthogonality statements concern controlled refitting paths rather than a general efficiency theorem. The transferable object is the algebra itself: typed residual ownership, ordered composition, and one auditable quotient boundary.

\subsubsection*{Reproducibility statement}
All architectural decisions, signal selections, and hyperparameters were frozen before the evaluation window; the paper reports the registered contrasts rather than post-hoc winners. Appendices~\ref{app:protocol}--\ref{app:audit-ledger} disclose the evaluation protocol, structural controls, and audit criteria. 

\subsubsection*{Ethics statement}
This work studies representation and architecture questions using historical market data. Reported returns are historical research measurements under a frozen protocol, not investment advice; realistic costs are included but capacity, impact, and implementation constraints of live trading are out of scope. The study uses no personal data.

\newpage
\bibliography{factor_fields}
\bibliographystyle{plainnat}

\clearpage
\appendix
\section*{Appendix}

\section{Proprietary signal specification}
\label{app:fieldselect}
\label{app:fields}

The empirical instance uses three preselected Fold fields and one fixed auxiliary interface. Their identities, transforms, dimensionality, screening rule, and candidate-level diagnostics are omitted because they encode commercially sensitive alpha research. All such choices were frozen before the evaluation window and held constant across every structural comparison. The disclosed evidence should therefore be read as an architecture study conditional on a fixed signal interface, not as a recipe for reproducing the underlying trading signals.

\section{Experimental protocol}
\label{app:protocol}

\begin{table}[H]
\caption{Frozen protocol constants.}
\label{tab:protocol}
\begin{center}
\scriptsize
\begin{tabular}{llll}
\toprule
Item & Value & Item & Value \\
\midrule
Market & China A-shares & Portfolio rule & long-only, drop bottom 20\% \\
Label & 5-day forward return & Phased holdings & 21 overlapping sleeves \\
Fold bins per axis & 10 (daily rank deciles) & Transaction cost & blended 10/15 bp \\
States per field & 100 & Time-decay half-life & 252 trading days \\
Strong fields & 3 (heterogeneous) & XGB learning rate / depth & 0.035 / 5 \\
Selection cutoff & 2022-12-31 & XGB max trees / early stop & 600 / 40 rounds (prior year) \\
Evaluation & 2023--2026, annual PIT & XGB min-child-weight / $\lambda$ & 2000 / 10 \\
Strict book & 3{,}666{,}920 rows, 844 days & Purge gap & $h{+}1$ dates \\
Median stocks/day & 4{,}380 & Bootstrap & 21-day blocks, 10{,}000 draws \\
Local expert interface & field coordinates + frozen auxiliaries & Available book & 3{,}690{,}036 rows (sensitivity) \\
\bottomrule
\end{tabular}
\end{center}
\end{table}

\section{Operator audit details}
\label{app:audit}

\begin{figure}[H]
\centering
\includegraphics[width=0.78\linewidth]{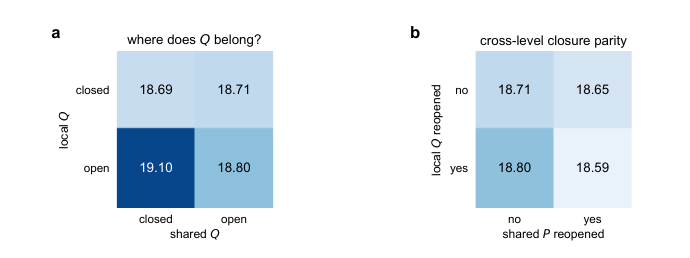}
\caption{\textbf{Capacity placement is an either/or.} Net return (\%) in two frozen $2{\times}2$ audits. \textbf{(a)}~A4: opening $Q$ locally is the best cell; adding shared $Q$ on top of local $Q$ \emph{reduces} net (interaction $-0.31\pp$, CI $[-0.35,-0.01]$). \textbf{(b)}~A3: reopening both cross-level edges is never the best cell.}
\label{fig:placement}
\end{figure}

\begin{table}[H]
\caption{Selected paired contrasts from the audit chain (strict book, pp, 10{,}000-draw block bootstrap).}
\label{tab:audit-inference}
\begin{center}
\scriptsize
\begin{tabular}{llccc}
\toprule
Audit & Contrast & $\Delta$ & 95\% CI & $\Pr[\Delta>0]$ \\
\midrule
A1 & shared-$Q$ closure over typed local-P & $+0.69$ & $[-0.30,1.72]$ & 0.91 \\
A1 & shared closure added over local-$Q$-only stage & $+3.43$ & $[0.66,6.06]$ & 0.99 \\
A2 & disagreement magnitude $D$ replacing $Q$ & $-0.09$ & $[-0.67,0.40]$ & 0.31 \\
A2 & directional contrasts $C_1,C_2$ beside $Q$ & $-0.14$ & $[-0.52,0.20]$ & 0.19 \\
A3 & shared-$P$ edge $\mid$ local-$Q$ open & $-0.21$ & $[-0.50,0.10]$ & 0.10 \\
A4 & shared-$Q$ effect $\mid$ local-$Q$ open & $-0.30$ & $[-0.73,0.15]$ & 0.10 \\
A4 & local-only $Q$ vs.\ predecessor (SA-LFSI) & $+0.39$ & $[-0.23,0.95]$ & 0.86 \\
A5 & self-state vs.\ permuted self-state & $-0.38$ & $[-0.73,0.00]$ & 0.03 \\
\bottomrule
\end{tabular}
\end{center}
\end{table}

\textbf{Capacity audit.} The matched direct control uses exactly the frozen FPRC-PQ total tree count in each deployment year---282, 228, 804, and 360 trees for 2023--2026, respectively---with no early stopping. It reaches 16.970\% net and Sharpe 1.963 versus 19.098\% and 2.089 for FPRC-PQ; all four yearly comparisons favor FPRC. The point is therefore where the same tree budget is seated, not merely how many trees are fit.

\section{Generic model-class and exploratory controls}
\label{app:model-controls}

\begin{table}[H]
\caption{\textbf{Matched structural controls}, full Run49 strict book (3{,}666{,}920 rows). $\Delta$ is control minus FPRC-PQ, so negative values favor FPRC; ``wins'' counts control years above FPRC. The ordinary direct row reuses Run101; the remaining controls are Run137.}
\label{tab:model-controls}
\begin{center}
\small
\setlength{\tabcolsep}{5pt}
\begin{tabular}{lrrrr}
\toprule
Model & Net\% & Sharpe & $\Delta$ net (pp) & Wins \\
\midrule
\textbf{FPRC-PQ} & \textbf{19.098} & \textbf{2.089} & --- & --- \\
Direct GBDT (Run101) & 17.014 & 1.882 & $-2.083$ & 0/4 \\
Matched-capacity Direct & 16.970 & 1.963 & $-2.128$ & 0/4 \\
Field concat.\ + unified residual & 16.824 & 1.877 & $-2.273$ & 0/4 \\
Identity-free two-stage boosting & 17.408 & 1.931 & $-1.690$ & 0/4 \\
Pairwise-only HGB (GA$^2$M-class) & 16.681 & 1.939 & $-2.417$ & 0/4 \\
\bottomrule
\end{tabular}
\end{center}
\end{table}

The unified-residual comparison is the cleanest identity test: FPRC-PQ leads by $2.273\pp$ in 4/4 years, with a paired 21-day block-bootstrap interval of $[+0.123,+4.453]\pp$. The remaining FPRC-minus-control intervals are $[-0.670,+4.959]\pp$ for matched Direct, $[-0.351,+3.540]\pp$ for identity-free two-stage boosting, and $[-0.079,+4.861]\pp$ for pairwise HGB. The latter is deliberately described as GA$^2$M-class because it enforces pairwise-only interactions but is not an \texttt{interpret}/EBM implementation.

\begin{table}[H]
\caption{\textbf{Exploratory learned-aggregation control.} Only paired deltas within the common book are comparable; absolute returns or Sharpes must not be compared with Table~\ref{tab:model-controls}. The learned-MoE book has 2{,}628{,}660 rows. Deltas and intervals are in basis points.}
\label{tab:exploratory-controls}
\begin{center}
\small
\setlength{\tabcolsep}{4pt}
\begin{tabular}{llrrr}
\toprule
Family & Paired contrast & $\Delta$ net (bp) & Wins & 95\% CI (bp) \\
\midrule
Learned MoE & gate only $-$ equal mean + closer & $-59.3$ & 2/4 & $[-196.8,+86.2]$ \\
Learned MoE & gate + fresh closer $-$ equal mean + closer & $+12.8$ & 2/4 & $[-79.6,+123.0]$ \\
\bottomrule
\end{tabular}
\end{center}
\end{table}

\textbf{Learned aggregation.} The three typed local experts are frozen, and only annual PIT affine-softmax weights are learned. Gate-only is 59.3\,bp below the frozen equal-mean-plus-closer system. In the mechanism-matched full-interface comparison, restoring a fresh parents-only closer leaves the learned gate only $+12.8$\,bp ahead, with 2/4 winning years and an interval spanning zero. Learned weighting therefore does not stably improve the unit interface. Because this audit requires the narrower frozen-expert coverage intersection, its absolute performance cannot be compared with the full main table.

\section{Operator-composition diagnostics: ordering, coefficients, and residual closure}
\label{app:ordering}

Fold's projection view and FPRC's ordered routing motivate three structural predictions that we probed \emph{post hoc} on the frozen 2023--2026 book. These are finite-capacity diagnostics on the already-explored evaluation window, \emph{not} preregistered promotions; we report them as structural evidence for the composition, using the same paired 21-day block-bootstrap machinery as the main text.

\textbf{Ordering (non-commutativity).} Reversing the two stages---closing on the pooled fields first and applying the typed local relaxation only afterward ($\Cop$ before $\Rop$)---moves net return by $-0.12\pp$ versus the frozen local-before-shared chain (CI $[-1.07,1.02]$). The sign matches the ordered-closure reading (each operator should fit only the residual its predecessor leaves); the interval is wide, but taken with the \emph{significant} A5 rejection of letting the closer re-read local state, the composition's asymmetry is directional-to-significant and never neutral. The operators do not commute.

\textbf{Unit coefficients sit at a flat optimum.} The fixed identity-erasure mean (D2) gives every field an equal offset coefficient. Relaxing this buys nothing measurable and can only add estimation variance. An annually-refit \emph{common} closed-form scaling $\hat\alpha^{*}_{Y}$---whose fitted value wanders over $0.45$--$1.15$ across deployment years, so it is emphatically \emph{not} pinned to one---moves net return by $-0.01\pp$ (CI $[-0.21,0.18]$, $\Pr[\Delta>0]=0.46$). A two-coefficient analytic reweighting of the local and shared increments moves it by $+0.05\pp$ (CI $[-0.66,0.84]$, $\Pr=0.60$), and its per-year solve is itself ill-conditioned. Conversely, doubling the anchor to a matched $\alpha{=}2$ offset \emph{harms} all three chains ($-0.56$ to $-0.77\pp$, higher turnover and cost). The unit interface is therefore statistically indistinguishable from the analytically-tuned coefficients while carrying none of their variance, and strictly better than over-anchoring---the fixed point in \emph{performance} is flat around unit weights even though the coefficient itself is not identified at exactly one.

\textbf{Operational residual closure.} We fit diagnostic probes to the final residual $r-\widehat{S}$ over 2024--2026 (the first window with enough nested out-of-fold history). Neither a typed-local probe (three field-identity learners, equal-weighted) nor a shared parents-only probe recovers any out-of-sample signal: both have $R^{2}_{\mathrm{oos}}\le 0$ ($-2.5{\times}10^{-4}$ typed-local, $-1.3{\times}10^{-3}$ shared) with non-positive MSE balance. Within the tested local and shared hypothesis classes, the composition has exhausted the fields' recoverable structure---an \emph{operational} closure, not exact orthogonal completeness.

\section{Negative-control details}
\label{app:controls}

\begin{table}[H]
\caption{Preregistered generic meta-information controls versus frozen FPRC-PQ (strict book, pp).}
\label{tab:controls}
\begin{center}
\scriptsize
\setlength{\tabcolsep}{4pt}
\begin{tabular}{llcccc}
\toprule
Control & Placement & Net\% & $\Delta$ & 95\% CI & $\Pr[\Delta>0]$ \\
\midrule
Local alert $C^{\mathrm{alert}}$ & local & 18.88 & $-0.22$ & $[-0.51,0.07]$ & 0.07 \\
Expert-order entropy $H_{\mathrm{ord}}$ & shared & 18.77 & $-0.32$ & $[-0.71,0.05]$ & 0.05 \\
Field std & shared & 18.94 & $-0.16$ & $[-0.34,-0.02]$ & 0.01 \\
Field skew & shared & 18.98 & $-0.12$ & $[-0.22,-0.03]$ & 0.01 \\
Map drift $D_{\mathrm{map}}$ & shared & 17.97 & $-1.13$ & $[-1.93,-0.21]$ & 0.01 \\
Drift+std+skew & shared & 17.84 & $-1.25$ & $[-2.07,-0.37]$ & 0.003 \\
All metadata & local+shared & 17.76 & $-1.34$ & $[-2.14,-0.43]$ & 0.002 \\
DTS directed surprise & local & 19.03 & $-0.06$ & $[-0.35,0.19]$ & 0.31 \\
\bottomrule
\end{tabular}
\end{center}
\end{table}

The alert and drift channels are not information-free: $D_{\mathrm{map}}$ correlates $0.24$ with committee-level absolute error and $C^{\mathrm{alert}}_1$ positively with local absolute residuals. They fail because diagnosis and correction are different operations. Once field identities have been erased, a shared learner can use such metadata only for global recalibration; exposing it there also violates the fresh-residual boundary isolated by audit A5. The result supports the algebra's placement rule, not a general claim that metadata are uninformative.

\section{Yearly results and coverage sensitivity}
\label{app:yearly}

\begin{table}[H]
\caption{Yearly net return (\%), strict book. 2026 covers January--June (forward stub).}
\label{tab:yearly}
\begin{center}
\scriptsize
\begin{tabular}{lcccc}
\toprule
Model & 2023 & 2024 & 2025 & 2026 \\
\midrule
Mean of Fold fields & 2.70 & 7.22 & 4.83 & $-1.24$ \\
Shared correction only & 3.21 & 7.91 & 5.71 & 1.18 \\
Typed local-PQ only & 2.78 & 7.65 & 5.80 & 2.06 \\
Local-P $\to$ shared-Q & 3.26 & 7.89 & 5.81 & 1.75 \\
FPRC-PQ & 3.42 & 7.89 & 5.73 & 2.06 \\
\bottomrule
\end{tabular}
\end{center}
\end{table}

On the available book (3{,}690{,}036 rows), FPRC-PQ earns 19.24\% net. Every registered structural contrast keeps its sign and approximate magnitude, including self-state ($-0.44\pp$) and all-metadata ($-1.32\pp$). No conclusion about the residual algebra reverses between the strict and available books.

\section{Reproduction and PIT audit ledger}
\label{app:audit-ledger}

Every run re-derives its frozen predecessor signal before reporting: the maximum absolute difference between rebuilt and frozen signals is $0.0$ in all audits. Native-missing gates are preserved (features may be missing; the sample gate never changes), the field tables and $D_{\mathrm{map}}$ observe labels only through $t-h-1$, and the annual purge covers the five-day forward label. $Q$-axis coverage is 93.9\% of rows. Total wall clock per audit is 2.5--7.4 minutes on four GPU workers.

\section{Use of large language models}
An LLM assisted with manuscript drafting (prose, LaTeX, and figure scripting) under author direction. All experimental designs, frozen protocols, and numerical results were produced by the authors' pipeline; all numbers in the paper are copied from the frozen result ledgers, and the authors verified all claims.

\end{document}